# On Feature based Delaunay Triangulation for Palmprint Recognition

*,1Zanobya N. Khan, 2Rashid Jalal Qureshi, and 3Jamil Ahmad

*,1Department of CS & IT, Sarhad University of Science & IT, Pakistan,
zanoby.nisar@gmail.com

2,3Department of Computer Science, Islamia College University, Pakistan,
rashid.jalal@gmail.com, jamil.ahmad@icp.edu.pk

*Corresponding Author

*Abstract*

*Authentication of individuals via palmprint based biometric system is becoming very popular due to its reliability as it contains unique and stable features. In this paper, we present a novel approach for palmprint recognition and its representation. To extract the palm lines, local thresholding technique Niblack binarization algorithm is adopted. The endpoints of these lines are determined and a connection is created among them using the Delaunay triangulation thereby generating a distinct topological structure of each palmprint. Next, we extract different geometric as well as quantitative features from the triangles of the Delaunay triangulation that assist in identifying different individuals. To ensure that the proposed approach is invariant to rotation and scaling, features were made relative to topological and geometrical structure of the palmprint. The similarity of the two palmprints is computed using the weighted sum approach and compared with the k-nearest neighbor. The experimental results obtained reflect the effectiveness of the proposed approach to discriminate between different palmprint images and thus achieved a recognition rate of 90% over large databases.*

**Keyword**s: delaunay triangulation, niblack binarization method, sorensen distance

## 1. Introduction

Biometrics is used to recognize or verify humans on the basis of their physical or behavioral characteristics [1]. For the security of access controls several features such as face, iris, fingerprint, hand geometry, palmprint, signature, etc, have been suggested. Most of the research in biometrics has been focused on fingerprint and face. However, due to factors such as aging, facial expression and the environment in which images are captured makes face based personal identification less reliable [2]. Fingerprint-based identification has been effectively used in various applications. Nevertheless, those of elderly people and manual labors are greatly worn down resulting in unclear images [3]. Therefore, it becomes difficult to acquire minutiae features of such class of people. Palmprint, as compared to other modalities has received considerable research interest because of many attractive advantages such as low resolution imaging, inexpensive capturing devices, easy self-positioning, public acceptability, high stability of features and is less vulnerable to noise interference during extraction [4-7]. Palmprint contains a variety of discriminative features such as texture, appearance, geometric features, principal lines, wrinkles, minutiae and delta points.

Palm lines, comprising both principal lines and wrinkles play a vital role to discriminate between different palmprints. Therefore, palm lines are atoned as one of the most significant features in automated palmprint recognition. Hand geometry features (such as palm length, width, finger length etc.) are not unique and can be susceptible to lighting conditions, contrast and position changes. Therefore, these systems are less accurate in a large database [8]. Coding based methods which includes the response of a bank of filters applied at different orientation results is successful representation of the palmprint images. However, in these methods, both the bitwise coding features and the integer representation produce information at pixel level. The pixel level information results in limit discriminability and sensitivity to the variation of translation and rotation because of imperfect preprocessing [5, 9].

In this article, we adopt a novel approach for palmprint recognition. Figure 1 summarizes our palmprint recognition approach. As such, for the extraction of the palm lines, we have explored the Niblack binarization [18] algorithm. Then through the endpoints of the palm lines, a connection is created using the DT, thereby generating a distinct topological structure of each palmprint. We exploit the geometrical as well as quantitative features from the triangles of the Delaunay triangulation that assist in identifying different individuals. To ensure that the proposed approach is invariant to rotation and scaling, relative features are adopted for the topological and geometric structure of the palm. The similarity of the two palmprints is computed using the weighted sum approach. The experimental results obtained show the viability and effectiveness of the proposed approach and adds to the discriminating power of our palmprint recognition approach. Overall, we achieve an acceptable recognition rate of 97% using the state-of-art databases.

## 2. Related Works

In [10], Patprapa used an average filter and magnitude gradient to extract principal lines. Morphological operators are then applied to smooth the contours and fill small holes. However, this method failed to achieve high accuracy. Wong [11] employs Sobel masks of two sizes (3x3 and 5x5) to extract principal lines and used Hamming distance as a classifier. Akinile [12] also employ Sobel masks and morphological operators to extract principal lines. Distances from one endpoint to another and from point of intersection to endpoints are calculated and transform into frequency domain using Discrete Fourier Transform (DFT). Zhu [13] propose another algorithm which use Canny edge detector, locally self-adaptive threshold binarization and morphological operators to extract principal lines. The feature vector is compared using fuzzy logic. Bouchemha [14] extracted principal lines using line directional mask. Radon Transform and adaptive thresholding is then applied to extract maximum coefficients along each projection from which centroids are extracted. These centroids are connected to each other by means of Delaunay triangulation and matched using Hausdorff distance. The Delaunay triangulation has been used widely for fingerprint and face recognition. G. Behis [15] proposed fingerprint identification system using DT for indexing. This reduced the memory requirements without compromising the recognition accuracy and persevered similarity without restoring to high dimensional indexing. In [16], Ning Liu used DT for fingerprint matching and developed a matching algorithm based on DT Net to find reference minutiae pairs.

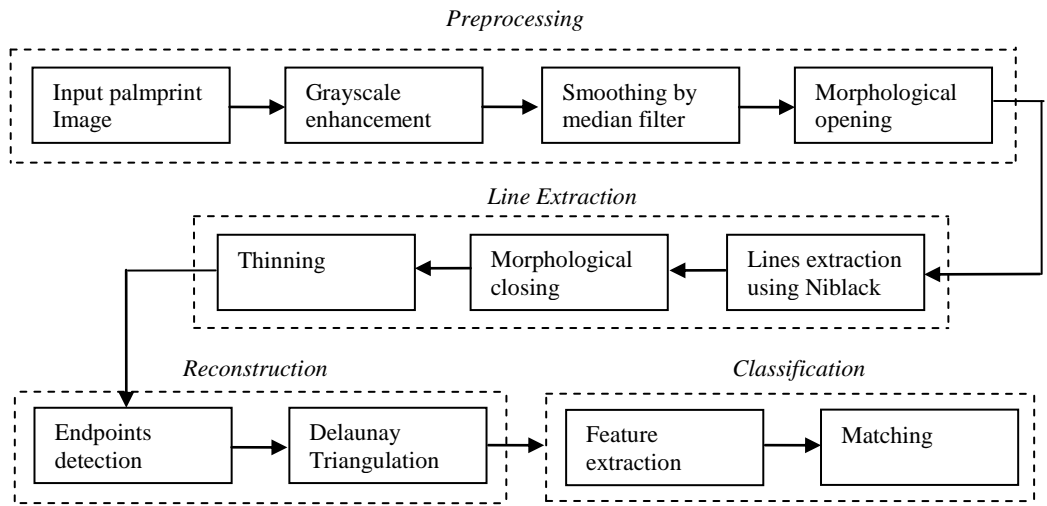

**Figure 1.** Flow diagram of the proposed approach

## 3. Feature Representation

The line features in the palmprint images which include both principal lines and wrinkles are extracted using Niblack binarization method as in [17]. From the extracted lines features, endpoints are then detected for all the lines. The endpoints detected from the lines of the palmprint images depict scattered data points. Therefore, DT is an effective tool for the representation of these scattered data set or points. The DT of each palmprint can be created by connecting the coordinates of each of the endpoint to its nearest neighbor such that the circumcircle associated with each triangle does not contain a point in its interior. Thus, each palmprint is represented by its unique topological structure as shown in Figure 2.

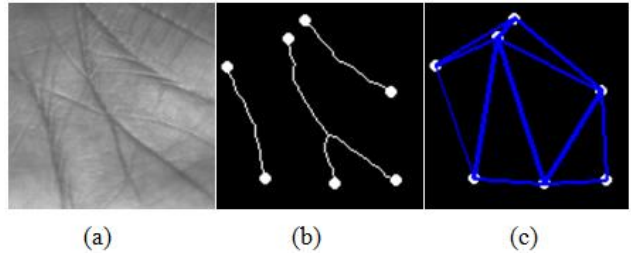

**Figure 2.** (a) Palmprint (b) Lines with endpoints (c) DT of corresponding endpoints

## 4. Feature Extraction from DT

After obtaining the unique topological structure of palmprint using DT, we will now extract feature from these representations. The feature selection is done keeping in view their discrimination power, and ease of computation. To obtain rotation, translation and scale invariant properties, certain features are made relative e.g., relative lengths of edges of a triangles, relative surface area of the triangles etc. The quantitative features are acquired by analyzing the DT quantitatively such as the number of triangles in the triangulation. It has been observed that changes in palm line produce different number of triangles in a DT. Therefore, it can be used as a filter for clustering similar palmprint images quickly. Line structure of some palmprint images may be similar as such their resultant DT may have a similar shape. But it is certain that relative length (RL) of the edges of all these polygons and their angles cannot be same. As such, range features serve as major decisive features to discriminate palmprint belonging to different classes [18]. The range features are explained in the following sections. The range features are explained in the following section.

### 4.1. Relative Length

The line segment is defined with respect to its endpoints in a coordinate system. The vertices of the triangle correspond to the endpoints and the line segment (i.e. edge) represents the connection between these endpoints. Therefore, the coordinates of the neighboring endpoints can be used to compute the length of each edge of the triangle using the Euclidean distance described as

$$\ell_i = \sqrt{\sum_{i=1}^{N}\left(\left(x_2(i) - x_1(i)\right)^2 + \left(y_2(i) - y_1(i)\right)^2\right)} \qquad (1)$$

We used relative length (RL) instead of length in pixels to allow the proposed algorithm to be scale invariant. We find the lengths of each edge, store them and sort them in a list, next we select the longest edge and consider it as $\ell_{max}$. The relative length $DL_i$ of the $i^{th}$ edge of each of the triangle can be computed as a ratio between its length ($\ell_i$) and the length of the longest edge ($\ell_{max}$) found in a particular DT can be defined as:

$$DL_i = \ell_i / \ell_{max} \qquad (2)$$

The relative length (DL), which is computed for each edge of the DT, is discretized into five classes at an interval of 0.2 to enable incorrect matching of the occurrences of the same class that have small distortion [18].

$DL_1$ = edges included in the range of $0.2 \leq DL \geq 0.0$

$DL_2$ = edges included in the range of $0.4 \leq DL > 0.2$

$DL_3$ = edges included in the range of $0.6 \leq DL > 0.4$

$DL_4$ = edges included in the range of $0.8 \leq DL > 0.6$

$DL_5$ = edges included in the range of $1.0 \leq DL > 0.8$

## 4.2. Relative Area

The area of each of the independent triangle can be computed easily using the information available at the vertices of the DTs. The area of a triangle can be defined as half the product of base and its corresponding height. The parameters of the area can be calculated using the Pythagoras theorem. However, relative areas ($DA_i$) will be considered to ensure scale invariant property of the proposed algorithm. It can be defined as

$$DA_i = A_i / A_{max} \qquad (3)$$

The relative areas ($DA_i$) for each of the DT is also normalized and divided into five classes of interval 0.2 as,

$DA_1$ = triangles with areas between $0.2 \leq DA_i \geq 0.0$

$DA_2$ = triangles with areas between $0.4 \leq DA_i > 0.2$

$DA_3$ = triangles with areas between $0.6 \leq DA_i > 0.4$

$DA_4$ = triangles with areas between $0.8 \leq DA_i > 0.6$

$DA_5$ = triangles with areas between $1.0 \leq DA_i > 0.8$

## 4.3. Angle

All printed Angle can also serve as an important discriminant feature which will be helpful in classifying different palmprints. Given the information at the vertices of the triangle, the angle ($\varphi$, [0, $\pi$]) with horizontal axis can be determined using the following equation,

$$D\theta_i = \tan^{-1} \frac{(y_2(i) - y_1(i))^2}{(x_2(i) - x_1(i))^2} \qquad (4)$$

Similarly, angle of each edge of the triangle that has values between 0° and 180° is divided into 6 classes. These features include number of edges with angle in intervals of 30°

$D\Theta_1$ = edges angle in range of 0° and 30°

$D\Theta_2$ = edges angle in range of 30° and 60°

$D\Theta_3$ = edges angle in range of 60° and 90°

$D\Theta_4$ = edges angle in range of 90° and 120°

$D\Theta_5$ = edges angle in range of 120° and 150°

$D\Theta_6$ = edges angle in range of 150° and 180°

## 4.4. Relative Incenters

Similarly, angle of each edge of the triangle that has values between 0° and 180° is divided into 6 classes. These features include number of edges with angle in intervals of $30^o$

$$DC_i = c_i / c_{max} \qquad (5)$$

Like relative length and area, the incenter for each triangle in the DT, is also categorized into five classes at an interval of 0.2.

$DC_1$= triangles incenters between $0.2 \leq DC_i \geq 0.0$

$DC_2$ = triangles incenters between $0.4 \leq DC_i > 0.2$

$DC_3$ = triangles incenters between $0.6 \leq DC_i > 0.4$

$DC_4$= triangles incenters between $0.8 \leq DC_i > 0.6$

$DC_5$= triangles incenters between $1.0 \leq DC_i > 0.8$

## 5. Result and Discussion

Figure 3 and Figure 4 shows the discrimination power of the features selected for the purpose of palmprint recognition. In Figure 3, the features extracted from the palmprints of the same person are shown. The lines in graphical representation coincide most of the time, showing a coherence behavior of the proposed method. On the contrary, the features of two different palmprints are plotted in Figure 4 shows the values of the features are relatively different.

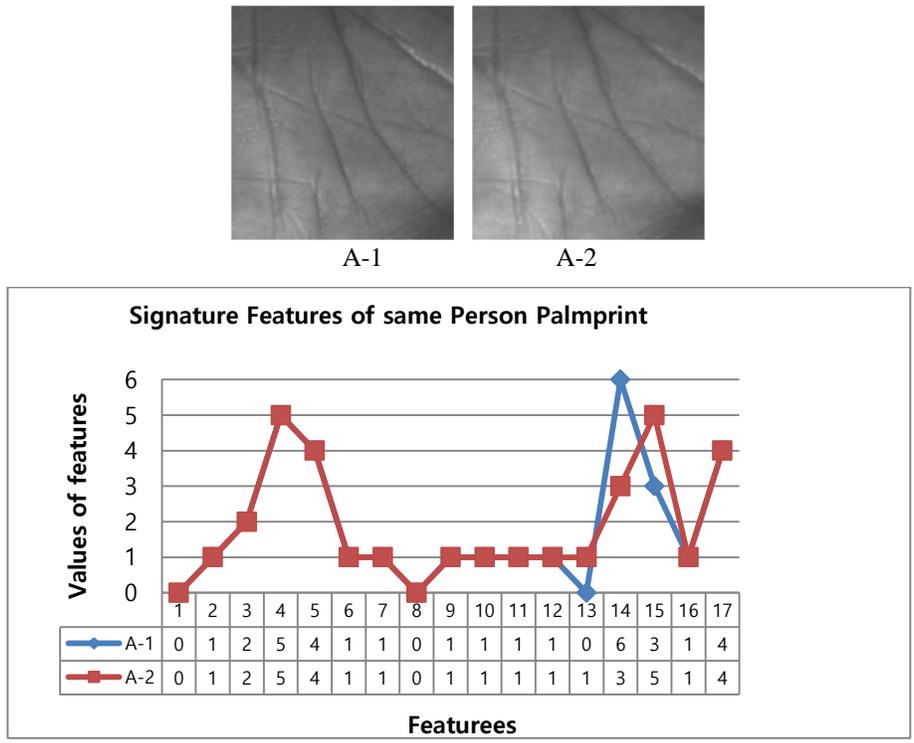

**Figure 3.** Feature vector values associated with same palmprint

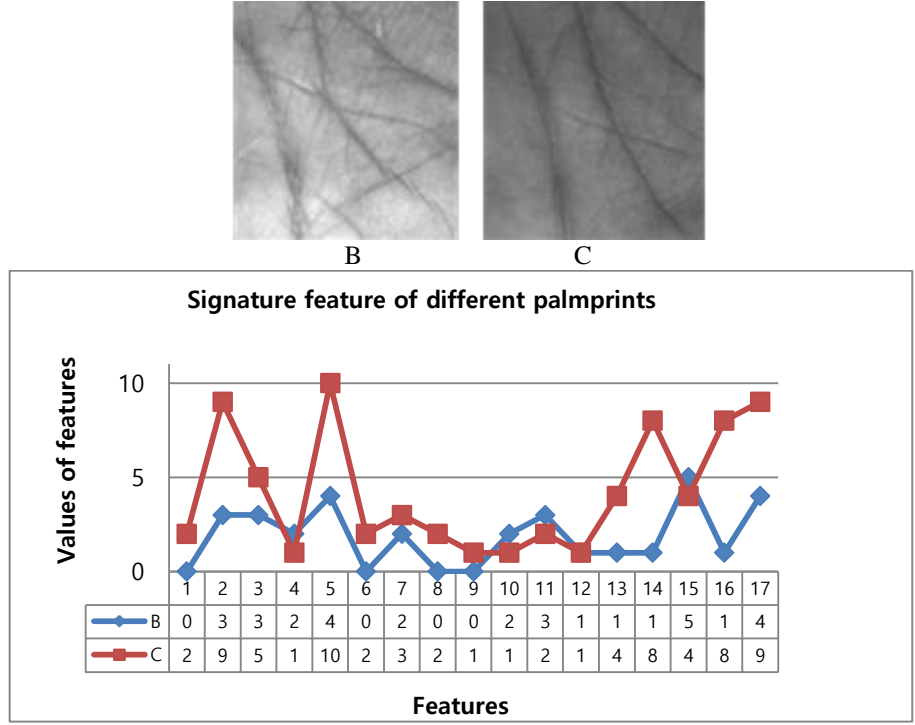

**Figure 4.** Comparison of feature vector values associated with different palmprints

## 5.1. Palmprint Matching

In the proposed approach, we used weighted sum strategy and K-nearest neighbor [19] to measure the similarity between palmprints. We have used weighted sum as some features serves to be more discriminant as compared to others and they have been assigned greater weights. Let $F_i$ and $F_j$ be two feature vectors then,

$$S(F_i, F_j) = \alpha \mathcal{D}L + \beta \mathcal{D}A + \gamma \mathcal{D}\varphi + \delta \mathcal{D}C \qquad (6)$$

where α, β, γ, and δ are the weights and $\mathcal{D}$V, $\mathcal{D}$A, $\mathcal{D}$φ and $\mathcal{D}\mathcal{J}$ are the distances returning a value between 0 and 1. We have used Sorensen [20] as a distance metric. The distance $d_{ij}$ between feature vectors $F_i$ and $F_j$ can be described as:

$$d_{ij} = \frac{\sum_{k=1}^{n}|x_{ik} - x_{jk}|}{\sum_{k=1}^{n}(x_{ik}) + \sum_{k=1}^{n}(x_{jk})} \qquad (7)$$

Where $x_{ik}$ and $x_{jk}$ are the $k^{th}$ elements of the of the feature vector $F_i$ and $F_j$

$$F_i = [x_{i1}, x_{i2}, \dots x_{ik}] \qquad (8)$$
$$F_j = [x_{j1}, x_{j2}, \dots x_{jk}]$$

Similarly, $\mathcal{D}$L, $\mathcal{D}$A, $\mathcal{D}$φ and $\mathcal{D}C$ correspond to the distance calculated between relative length, relative area, angle and incenter extracted from Delaunay triangulation of the two palmprints. The closer the value to zero, greater the similarity ratio between two palmprints whereas a high value defines dissimilarity. Moreover, the matching value 0 specifies a perfect match.

**5.2. Database Setup**

The performance of our proposed approach was evaluated by carrying out a series of experiment on the palmprint database (2D Palmprint ROI) collected by the Polytechnic University Hong Kong palmprint database [21].This database contains 20 grayscale palmprint images of each individual that were collected in two sessions, where 10 samples were collected in the first session and remaining 10 in the second session respectively. A one month average time interval was kept between the two sessions. The size of each of the original is 128x128 after cropping. In our experiments, we set up two databases with varying sizes for DB1= 240 and DB2=300 belonging to 40 and 50 individuals respectively.

**5.3. Performance Evaluation**

To assess the performance of the proposed technique we adopted leave-one- out strategy. If the query palmprint image is matched with one of the image from the same person's palmprint image then it is termed as a correct matching. But if the test image matches a palmprint image of another individual then it is counted as an incorrect matching. Figure 5 shows the recognition rate. The proposed system achieved a recognition rate of 90% on DB2 (over a sample of 300 images selected from 50 classes) when experiments are conducted using weighted sum. However, for the same databases, K-nearest neighbor (K-NN) achieved a recognition rate of 80% and 82% respectively.

The greater value of the distance indicates that the features selected to represent the palmprints in the proposed approach can effectively discriminate between different classes of palmprint images

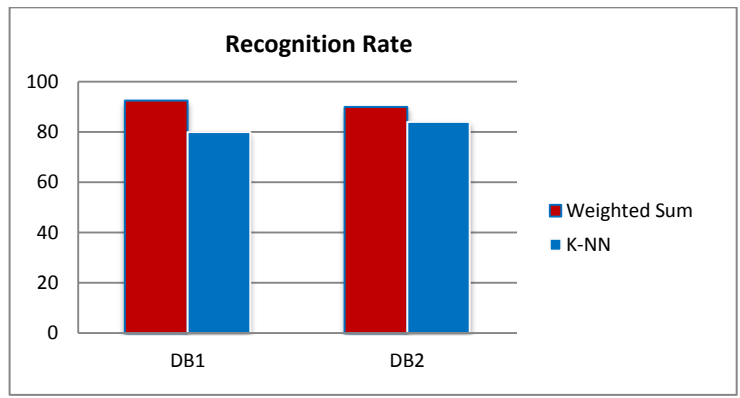

**Figure 5.** Recognition rate of the proposed approach

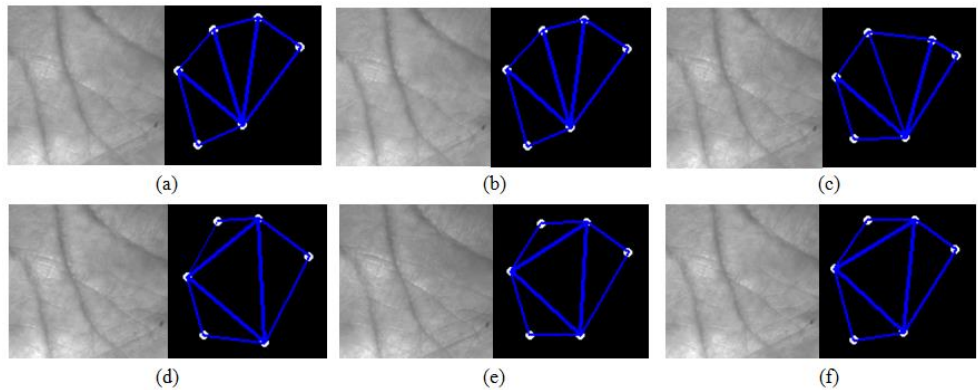

**Figure 6.** Instances of the same person palmprint images

The Table 1 shows the distances between different instances of the same person's palmprint images (Figure 6). The low values of distances show that the proposed system can accurately classify images. Similarly, Table 2 shows the distances among palmprint images that belong to different individuals (Figure 7). The greater value of the distance indicates that the features selected to represent the palmprints in the proposed approach can effectively discriminate between different classes of palmprint images.

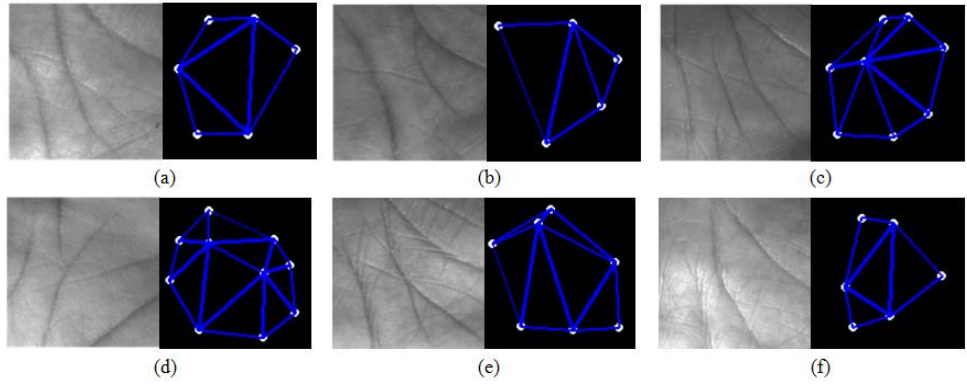

**Figure 7.** Palmprint images belonging to six different people

**Table 1.** Distances between palmprint images of Figure 5

| *Fig.5* | *a* | *b* | *c* | *d* | *e* | *f* |
|---|---|---|---|---|---|---|
| A | | 0.0052 | 0.0234 | 0.0494 | 0.0598 | 0.0651 |
| B | | | 0.0234 | 0.0494 | 0.0598 | 0.0703 |
| C | | | | 0.0520 | 0.0572 | 0.0520 |
| D | | | | | 0.0364 | 0.0572 |
| E | | | | | | 0.0312 |
| F | | | | | | |

**Table 2.** Distances between palmprint images of Figure 6

| *Fig.6* | *a* | *b* | *c* | *d* | *e* | *f* |
|---|---|---|---|---|---|---|
| A | | 0.3214 | 0.3560 | 0.5787 | 0.2250 | 0.2870 |
| B | | | 0.4500 | 0.6470 | 0.3333 | 0.4062 |
| C | | | | 0.4285 | 0.2628 | 0.3402 |
| D | | | | | 0.4375 | 0.4736 |
| E | | | | | | 0.2575 |
| F | | | | | | |

## 6. Runtime Analysis

In order to determine the computational efficiency of the proposed approach, it is compared with A. Bouchema et al. [14] approach. In [14], the authors extracted the principal lines using line detectors in four different orientations. Radon Transform and adaptive thresholding is then applied to extract maximum coefficients along each projection from which centroids are extracted. These centroids are connected to each other by means of DT. The total runtime required from line extraction to construction of DT is 1.78 seconds as depicted in Figure 8. However, from Figure 8 it can be seen that the execution time for the palm line extraction, their endpoints including preprocessing and construction of DT of the proposed approach (on average for a single image) is 0.42 seconds. Thus, it can be deduced from the experiment that our approach is less expensive computationally.

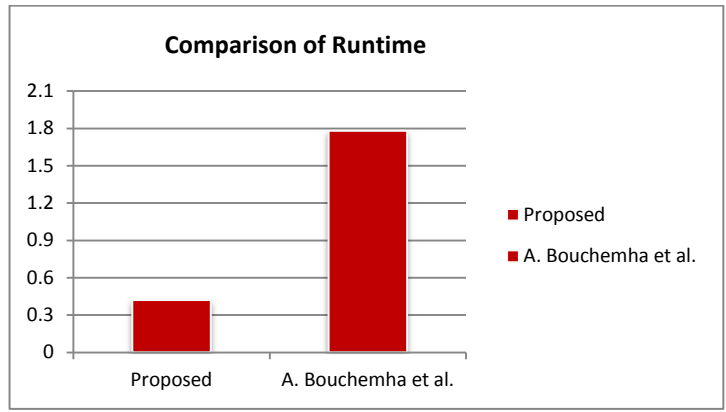

**Figure 8.** Runtime comparison of the proposed approach

## 7. Conclusion

Palm lines comprising both principal lines and wrinkles are one of the most important features for palmprint recognition systems. In this paper, we present a novel approach to extract line features and its representation. The use of relative features has made the proposed approach invariant to rotation, translation and scaling. The experimental results and the recognition rate demonstrate that our method is very effective.